\documentclass[10pt, a4paper, table]{article}

\usepackage{lrec-coling2024} % this is the new style
\usepackage{xcolor}
\usepackage{enumitem}
\usepackage{booktabs}
\usepackage{graphicx}
\usepackage{colortbl}
\usepackage[bottom]{footmisc}
\usepackage{dblfloatfix}
\usepackage{float}
\usepackage{subfigure}
\title{Specifying Genericity through Inclusiveness and Abstractness Continuous Scales} % proposta di sintesi

\name{Claudia Collacciani, Andrea Amelio Ravelli, Marianna Marcella Bolognesi} 

\address{Faculty of Modern Languages, Literatures and Cultures \\
        University of Bologna, Bologna, Italy \\
         \{claudia.collacciani2, andreaamelio.ravelli, m.bolognesi\}@unibo.it\\}

\abstract{
This paper introduces a novel annotation framework for the fine-grained modeling of Noun Phrases' (NPs) genericity in natural language. The framework is designed to be simple and intuitive, making it accessible to non-expert annotators and suitable for crowd-sourced tasks. Drawing from theoretical and cognitive literature on genericity, this framework is grounded in established linguistic theory. Through a pilot study, we created a small but crucial annotated dataset of 324 sentences, serving as a foundation for future research. To validate our approach, we conducted an evaluation comparing our continuous annotations with existing binary annotations on the same dataset, demonstrating the framework's effectiveness in capturing nuanced aspects of genericity. Our work offers a practical resource for linguists, providing a first annotated dataset and an annotation scheme designed to build real-language datasets that can be used in studies on the semantics of genericity, and NLP practitioners, contributing to the development of commonsense knowledge repositories valuable in enhancing various NLP applications.
 \\ \newline \Keywords{Collaborative Resource Construction and Crowdsourcing, Corpus, Semantics} }

\begin{document}

\maketitleabstract

\section{Introduction}

Language allows us to convey both information about particular individuals and situations, as in (1a), and generalizations about kinds, as in (1b). 

\begin{enumerate}[resume]
\item
    \begin{enumerate} [noitemsep]
        \item \textbf{The lion} escaped yesterday from the zoo.
        \item \textbf{The lion} is a predatory cat.
    \end{enumerate}
\end{enumerate}

The same noun phrase NP ("The lion" in the examples above) can be used in both interpretations. The syntactic form of the NP (definite, indefinite, plural) is not sufficient to disambiguate between the two meanings: the disambiguation is guided by the context in which the NP occurs \cite{Krifka1995-KRIGAI}.
This phenomenon can be found in every language \cite{behrens2005genericity} and in virtually all lexical items that can be employed in referring expressions (i.e., nouns).
Nevertheless, there is no explicit marker for generic NPs in natural languages \cite{dahl199512}: the NPs' genericity is determined by the meaning of the sentence as a whole.

Statements about kinds, such as (1b), defined \textit{generics}, can be seen as fundamental to human cognition, because they enable us to conceptualize properties associated with categories, structuring our perception of the world \cite{chatzigoga2019genericity}.
 \\ \newline
Existing annotation frameworks capture levels of genericity in linguistic expressions using discrete multi-class annotation schemes (\citealplanguageresource{ACE-2}; \citealplanguageresource{ACE2005}; \citealp{friedrich2015annotating}) or continuous multi-label systems \cite{govindarajan2019decomposing}. 
The use of such systems can be appropriate for cases such as those in (1a) and (1b), where the subject clearly refers respectively to a specific individual and to a category, and it is straightforward to assign each to the generic or non-generic meaning. However, the expression of genericity and its perception in speakers' minds is a very complex semantic aspect, which cannot be fully modeled by classifications of this kind. In fact, this type of annotations can denote a noun in context as generic or not (or even as both simultaneously, in the case of non-exclusive labels, such as those in \citealp{govindarajan2019decomposing}), but never specify \textit{how much} or \textit{in what way} it is generic.
In contrast, recent works in the cognitive domain tend to represent the expression and perception of genericity through continuous models \cite{tessler2019language}: as we will see, not all generics are the same.
\\ \newline
The most distinctive feature of generics is that they allow for exceptions \cite{Krifka1995-KRIGAI}, enabling speakers to interpret their quantification value in different ways relying on their world knowledge. This is shown from contrasts such as that between (2a) and (2b).

\begin{enumerate}[resume]
\item
    \begin{enumerate} [noitemsep]
        \item \textbf{Robins} are birds.
        \item \textbf{Robins} lay eggs.
    \end{enumerate}
\end{enumerate}

Both the statements in (2a) and (2b) are true and the NP subject is generic in both cases. However, in the first case it refers to all the individuals of the category (all robins are birds), while in the second it refers to only some of them (only adult females lay eggs).

Furthermore, as illustrated by the \textit{taxonomic reference} phenomenon \cite{carlson1995generic}, the same NP can refer not only to an individual and to a category, as in (3a) and (3b) respectively, but also to a subcategory, as in (3c).

\begin{enumerate}[resume]
\item
    \begin{enumerate} [noitemsep]
        \item \textbf{A whale} dove back into the ocean.
        \item \textbf{A whale} is a marine mammal.
        \item \textbf{A whale} which was recently put under protection is the blue whale.
    \end{enumerate}
\end{enumerate}

Another distinction is that between characterizing generics and direct kind predications \cite{carlson1977reference, sep-generics}. The first ones are statements such as (4a), which predicate applies to individual members of the kind; the second ones are sentences such as (4b), which predicate cannot refer to an individual, but only to a kind.

\begin{enumerate}[resume]
\item
    \begin{enumerate} [noitemsep]
        \item \textbf{Tigers} are striped.
        \item \textbf{Tigers} are widespread.
    \end{enumerate}
\end{enumerate}

It's worth noting that the indefinite singular form can be used in characterizing generics but not in direct kind predications: the sentence *\textit{A tiger is widespread} is not felicitous \footnote{Sentences such as this can be used only if the NP 'A tiger' is construed as a case of taxonomic reference, i.e., as referring to a species of tiger, thus still to the kind and not to the individual.}.

We also want to point out that the literature on genericity has hardly dealt explicitly with abstract nouns, tending to present the contrast between generic and non-generic meaning almost always through concrete nouns, as we also did in the examples given so far. This is probably the case because, intuitively, it appears that distinguishing between generic and non-generic meaning for abstract entities is less straightforward than for concrete ones.
\\ \newline
We propose a novel annotation framework to model the expression of NPs' genericity in a fine-grained manner, that is both grounded in linguistic theory and intuitive enough to be carried out by crowdsourcing. 

Our purpose is twofold. On one hand, we are interested in investigating if, from naive language users annotations, differences emerge that trace back to phenomena such as the ones above, observed in theoretical literature by experts (we carry out this investigation particularly in §5.2). 
On the other hand, we argue that annotations produced by our scheme can be useful to train systems to automatically identifying different levels of genericity; these, in turn, can be used to construct repositories of commonsense knowledge. In fact, given their nature, generic sentences are a powerful resource to retrieve common sense knowledge, exploitable to boost performance in various NLP applications, such as search, question-answering, and conversational bots. However, it is only recently that their usefulness in this regard has been proposed and demonstrated \cite{bhakthavatsalam2020genericskb, nguyen2023extracting}.
We propose that the possibility of identifying sentences not only as generic, but also being associated with values that model their semantics in a theoretical founded way, can help in constructing resources of this kind, that can be valuable for NLP applications, as well as for linguistics, providing real-language data to be used in studies on the semantics of genericity.

Our framework is designed to be language-independent.
In fact, as we have seen, the aspect of genericity is a very high-level semantic phenomenon, found in all languages yet in none explicitly marked.
\\ \newline
In this work, we focus exclusively on NPs' genericity, leaving aside that of predicates, which emerges in propositions that do not express specific episodes or isolated facts, but instead report a kind of a general property or a generalization over events, as in \textit{John drives to work every day} \cite{Krifka1995-KRIGAI}. By that, we do not mean to argue that the two are unrelated: indeed, very often the meaning of the predicate determines the interpretation of the noun to which it refers as generic or non-generic. However, we are interested in first proposing an annotation scheme that is ideally as informative as possible for nouns, since the sentences that refer to categories are the ones useful for the retrieval of commonsense knowledge.

\subsection*{Contributions}
\begin{itemize} 
    \item We propose a novel annotation framework for the annotation of NPs' genericity that: 1. is based on simple and intuitive evaluation mechanism and labels, so that it is suitable for crowd-sourced tasks; and 2. is grounded in the theoretical and cognitive literature on genericity (§3).
    \item We conduct a pilot study using this framework (§4), through which we produce a first small annotated dataset (324 sentences)\footnote{This dataset is not yet publicly released because it is currently being expanded and extended to other languages, with the objective to use it in a shared task in an upcoming evaluation campaigns. In due course, the dataset will be available at \url{https://osf.io/8w6u9/?view_only=9e9365d5bb8f4dba83b4081112e703ce}}.
    \item We evaluate our continuous annotations comparing them with pre-existing binary annotations on the same dataset (§5).
\end{itemize}

\section{Related work}

One of the first frameworks that explicitly aims at modelling semantic aspects of genericity was developed under the ACE-2 program (\citealplanguageresource{ACE-2}; \citealp{doddington-etal-2004-automatic}). This framework labels NPs as SPECIFIC when they refer to a precise member of the category and GENERIC when they refer to any member of the category. The ACE-2005 Multilingual Training Corpus \citelanguageresource{ACE2005} extends the annotation guidelines, adding two additional classes: negatively quantified entries (NEG) and underspecified entries (USP), where the referent is ambiguous between GENERIC and SPECIFIC.
The Situation Entities (SitEnt) framework (\citealp{friedrich2014situation}; \citealp{friedrich2015annotating}; \citealp{friedrich2016situation}) improves the previous approach in two ways: first, it annotates both NPs and entire clauses for genericity. Second, with regard to the NPs annotation, the SitEnt guidelines, based on semantic theory, improve and clarify  the ACE guidelines, in which the notions of genericity and specificity are conflated: the label GENERIC is applied when the subject of the clause refers to a kind \textit{or} to an arbitrary members of a kind, while the label NON-GENERIC is applied when it refers to a particular individual. 
%The guidelines make explicit that the label NON-GENERIC also includes cases of nonspecific reference if the reader can infer that the clause makes a statement about some particular individual (or group of individuals), even if the identity is unknown, as \textit{lion} in \textit{A lion must have eaten the rabbit}. In this way The SitEnt guidelines, based on semantic theory, improve and clarify  the ACE guidelines, in which the notions of genericity and specificity are conflated.
%However, despite the guidelines accuracy, a coding scheme based on a limited set of discrete labels can hardly fully capture the range of phenomena that characterize the expression of generalization and their perception in the speakers' minds. 
 
The problematic nature of multi-class approaches is first addressed by \citet{govindarajan2019decomposing}, who proposed a semantic framework in which expressions of generalization are captured in a continuous multi-label system. 
%This was accomplished by decomposing categories such as episodic, habitual, and generic into simple referential properties of predicates and their arguments. 
Their argument protocol involves the assignment of the labels PARTICULAR (\textit{instantiated individuals}), KIND (\textit{kinds of individuals}), and ABSTRACT (\textit{intangible}) to each noun in argument function, together with the indication of the annotator's confidence about the different referential properties.  
 \\ \newline
We agree with the problematization of previous frameworks advanced in \citet{govindarajan2019decomposing} and with the idea of a continuous type of annotation. However, we argue that also their framework falls short in modelling in an accurate way the gradual nature of the semantics of genericity, which, in our view, is a key aspect of it. In fact, the indication of annotators' confidence is not a direct measure of continuity between labels: their framework still assign labels to word occurrences, being ineffective in differentiate among different types of generics.

In our framework, we bring the two labels "kind"  and "particular" back onto a single semantic axis, treating them as the two poles of a continuum, rather than considering them separate classes or different properties. We  operationalize this dimension as \textbf{inclusiveness}. On the other hand, we take into account the semantic dimension of \textbf{abstractness}, which is also continuous. We argue, drawing on the descriptive and theoretical literature, that using both dimensions can help us model NPs genericity in a more informative way, as we will explain in the next section.

\section{Annotation Framework}
\begin{figure}[t!b]
    \centering
    \begin{subfigure}
         \centering
          \includegraphics[width=0.5\textwidth]{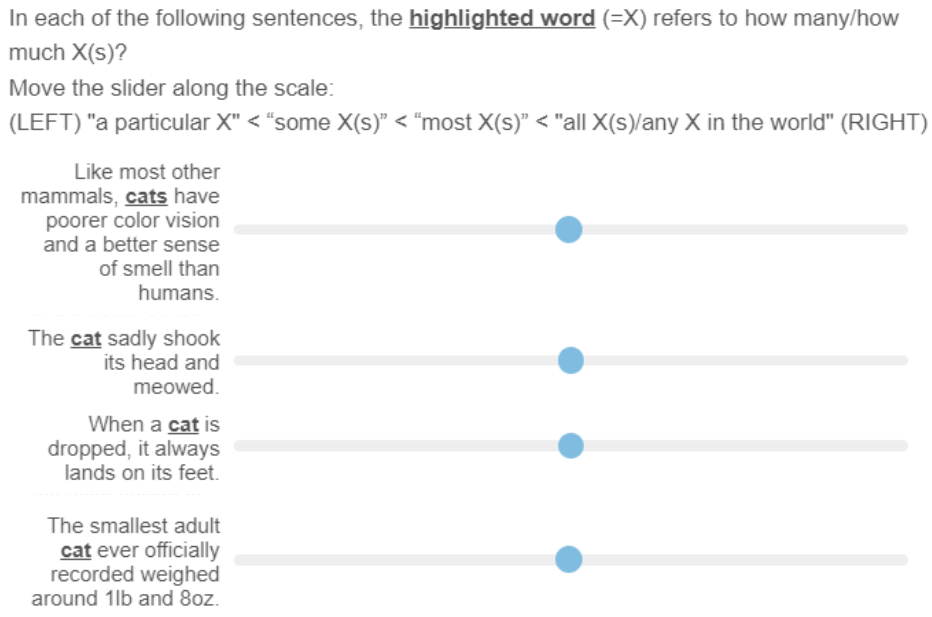}
    \end{subfigure}

    \begin{subfigure}
         \centering
          \includegraphics[width=0.5\textwidth]{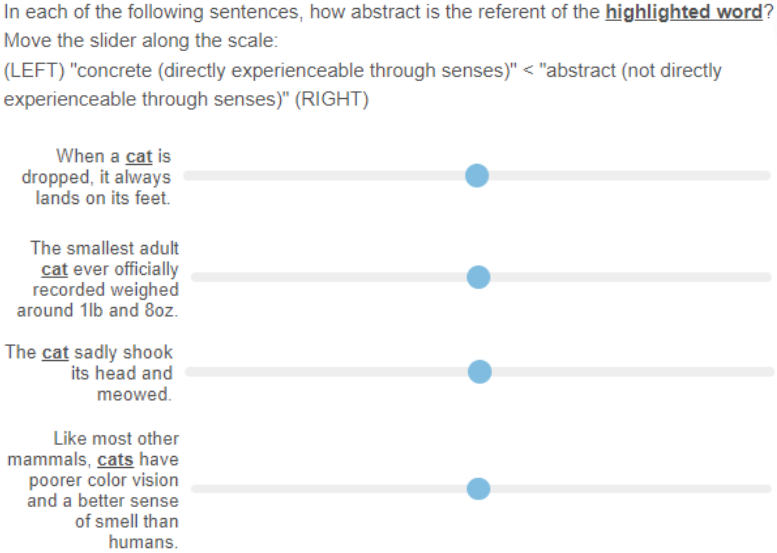}
    \end{subfigure}
    \caption{Examples of annotation interface for inclusiveness (top) and abstractness (bottom).}
    \label{fig:annotation_scheme}
\end{figure}

Through our annotation framework, we propose to model genericity through (i) two different semantic dimensions; (ii) continuous evaluations. 
 \\ \newline
The use of the two dimensions that we label inclusiveness and abstractness draws on the theoretical literature on genericity. 
As \citet{chatzigoga2019genericity} points out, we can divide theoretical accounts of genericity into two broad categories: those that treat generics as quantificational, for which generics quantify over members of the kind; and those that do not, for which generics are seen to predicate a property directly of the kind itself. These two views are the same that \citet{tessler2019language} call statistical and conceptual views of generic language. Through inclusiveness, we aim to grasp the quantificational, or statistical, aspect of genericity; through abstractness, we aim to grasp the non-quantificational, or conceptual, aspect of genericity. 
\\ \newline
As for the use of continuous evaluations, we argue that it is the best way to model a range of phenomena that are impossible to capture through a discrete label system, such as those mentioned in §1.
%A linguistic phenomenon can be annotated in various forms. For example, consider the case of the sentiment that characterises a linguistic expression. The first type of annotation we can apply is discrete, involving the assignment of a class from a binary set: NEGATIVE and POSITIVE. First used in psychology, but nowadays common in many other disciplines, graduated scale (i.e., Likert scales) are a finer way to tackle the annotation problem, where intermediate levels are placed between the (absolutely) negative and (absolutely) positive extremes, with increasing granularity as the scale extends. Finally, we can consider the phenomenon as continuous and, therefore, finely rate it using continuous scales. 
The most commonly used method for capturing fine semantic properties is certainly the use of Likert scales, typically with 5, 7 or even 9 points \cite{brysbaert2014concreteness, concretext2020task, montefinese2014adaptation}; however, continuous scales have many advantages over those, as remarked in many studies \cite{champney1941measurement, svensson2000comparison, belz2011discrete, ethayarajh2022authenticity}. From a psychological perspective, continuous scales are generally preferred by raters, as they do not need to discretise their choices. Another big advantage of continuous scales against Likert scales is that they avoid ordinal-cardinal conflation and potentially biased estimation \cite{ethayarajh2022authenticity}. Continuous scales in Natural Language Processing (NLP) have been mainly used to evaluate the quality of the output of Natural Language Generation (NLG) systems \cite{gatt2009tuna, bojar2017findings}. Beltz and Kow \citet{belz2011discrete} directly compare the two scaling methodology (i.e. graduated scales vs. continuous scales) in the task of qualitative assessment of Natural Language Generation (NLG) systems, concluding that the two metholodogies are interchangable and stable in the resulting annotation.
We argue that continuous scales are also suitable for the annotation of high-level semantic properties such as our dimensions. With our pilot study, we implicitly test also this claim: obtaining good reliability values confirms the suitability of the method.
 \\ \newline
We present the annotators with continuous sliders\footnote{From which values ranging from 0 to 1 will then be extracted.}, on which to evaluate the two dimensions, each in a separate task.
As shown in Figure \ref{fig:annotation_scheme}, our protocol proposes to the annotators groups of sentences (from a minimum of 4 to a maximum of 8), all containing the same noun, to be evaluated using the same scale.
We chose to display groups of sentences containing the same target noun in different contexts because genericity, and consequently the two continuous dimensions in which we unpack  it, are relational properties. A word can be considered more generic or less generic \textit{compared to} a related word, or to the same word used in a different context. To carry on the annotation in the correct way, annotators need to mentally put into relation the referent of the NP in each context to all the other existing referents denoted by the same word. Presenting sentences in groups of this type is a way to simplify this process for annotators, given the complex nature of the variables.

The complete instructions provided to the annotators are available in the OSF repository of this study\footnote{\url{https://osf.io/8w6u9/?view_only=9e9365d5bb8f4dba83b4081112e703ce}}.

\paragraph{Inclusiveness}
By inclusiveness, we refer to the quantificational aspect of the semantics of an NP, that is, how many members of the category the NP refers to. The number of members of the category who possess the property is called \textit{prevalence} in literature on genericity; we use the term "inclusiveness" to refer not only to generic NP, but also non-generic ones. We draw inspiration from \citet{herbelot2010annotating}. They point out that many NPs are not explicitly quantified; still, humans are able to give them quantificational interpretations in context (e.g., Cats are mammals = \textit{All} cats...; Water dripped through the ceiling = \textit{Some} water...). The authors label this phenomenon 'underquantification', and argue that it also applies to generic NPs. 

Their annotation scheme assumes a three-fold quantificational space, corresponding to the quantifiers \textit{some}, \textit{most} and \textit{all}, in addition to \textit{one}, for singular and unique entities. Following \citet{tessler2019language}, who propose a generalization of underquantification to a continuous interval of possible meanings,we adopt a continuous quantificational space, in which quantifiers are used as a reference (Figure \ref{fig:annotation_scheme}, top).

We argue that the use of a quantification continuum makes it fairly intuitive to account for the inclusiveness of NPs. From our perspective, such a scheme is felicitously applicable to nouns referring to abstract entities as well. In fact, according to Moltmann \shortcite{moltmann2004properties, moltmann2013abstract}, bare singular abstract nouns, such as that in (5a), are kinds, in that they denotes \textit{kinds of tropes}, while ‘tropes’ are specific instances of property attribution, such as the same noun in (5b).
\begin{enumerate}[resume]
\item
    \begin{enumerate} [noitemsep]
        \item \textbf{Ordinariness} is boring.
        \item John's \textbf{ordinariness} is disarming.
    \end{enumerate}
\end{enumerate}

Such a contrast is not straightforward to capture through the assignment of labels such as generic/non-generic or kind/particular, but we propose that can be more easily detected by the use of a quantification continuum.

\citet{herbelot2010annotating} also claim that while generics can always be quantified, their semantics may involve more than quantification. This seems to be especially true for direct kind predications.
Herbelot and Copestake \shortcite{herbelot2010annotating,herbelot2011formalising}, following the formal accounts of \citet{chierchia1998reference} and \citet{krifka2003bare}, argue that also generics of this type do not preclude quantification, but that in certain cases a description of the NP both as a quantified entity and a kind is nevertheless desirable. This position is close to that of \citet{tessler2019language}, who propose a semantic core based on prevalence, but argue that prevalence is not enough to capture subtle sensitivities to context that generic language shows. 
This is precisely why we incorporated another dimension into our framework, to try to account for the multidimensional nature of genericity.

\paragraph{Abstractness}
Through abstractness, we aim to capture another aspect of the semantics of an NP, that is, to what extent the NP denotes a referent that can be experienced through sensory modalities (Figure \ref{fig:annotation_scheme}, bottom).
The first reason we incorporated this dimension is that there is a possibility that concrete concepts' kinds are perceived to be more abstract than concrete concepts' instances (\citealp{pelletier2009kinds};\citealp{zamparelli2020countability}).
This seems to apply primarily to direct kind predications, since in such cases the predicate is not applicable to the individual concrete objects, but only to the kind.

Moreover, there are many cases of fine-grained polysemy in which the same word has multiple senses that are closely related but differ in abstractness. Consider the case of \textit{church}, which can refer to the building or the social group, depending on the context; or \textit{book}, which can refer to the physical object or the content. Similarly, there are abstract nouns whose countability status can shift depending on the context in which they occur, and is determined by their designation or construction in a particular occurrence \cite{grimm2014individuating}. 
These nouns (e.g., \textit{agreement} (i.e. a state of concord) vs. the recently signed \textit{agreements}), afford  'elastic' changes between their mass and count use. We argue that they afford also changes in their degree of abstractness \cite{zamparelli2020countability}.

We claim that the aspect of abstractness is somehow conflated into a coarse-grained classification such as the GENERIC/NON-GENERIC one. 
We therefore separate inclusiveness and abstractness while restoring their continuous nature. This approach allows us to model the semantics of genericity in a finer-grained manner, which arguably better captures the complexity found in human interpretations.

\section{Pilot Study for Framework Validation}
We conducted a pilot study to validate our annotation framework from two different points of view: (i) whether the annotators show fair agreement on the evaluations; (ii) whether our framework, based on annotations provided by crowd workers, subsumes the binary GENERIC/NON-GENERIC distinction annotated by experts. This second point (which we will discuss in §5) is necessary to ensure that the evaluations collected actually reflect the semantic aspect of genericity, on which the previous annotations are grounded; at the same time, it aims to show that our scheme adds information with respect to them.

\paragraph{Dataset}
We used a sample of 324 sentences extracted from the SitEnt dataset (\citealp{friedrich2014situation}; \citealp{friedrich2015annotating}; \citealp{friedrich2016situation}). On this sample, we annotated the \texttt{mainReferent} (as defined by SitEnt) of each sentence, that was already labeled as GENERIC or NON-GENERIC.
The unique nouns in the dataset are 60, and each occurs in a minimum of 4 to a maximum of 8 different sentences, which are presented to the annotators in groups (Figure \ref{fig:annotation_scheme}). We tried to keep the sample as balanced as possible between GENERIC and NON-GENERIC NPs, with 165 NON-GENERIC and 159 GENERIC target nouns; moreover, each unique noun occurs at least once as GENERIC and once as NON-GENERIC. 

Furthermore, we kept the ratio between concrete and abstract target nouns constant with respect to that of the whole SitEnt dataset (70-30\% ca.). We associated with each \texttt{mainReferent} the concreteness value for the corresponding lemma retrieved from \citet{brysbaert2014concreteness} and considered as concrete those nouns associated with scores greater than 3 and as abstract those nouns associated with scores lower than or equal to 3 (\citet{brysbaert2014concreteness}'s scale ranges from 1 to 5).

\paragraph{Annotators}
A total of 480 crowd workers, native speakers of English, were recruited as annotators through the Prolific platform; 240 annotated inclusiveness, 240 annotated abstractness. Sentence groups were presented in batches of 6, 8 or 10, with each target noun annotated by 30 annotators.
The SitEnt annotators were highly trained and were provided with the entire document containing the sentence to be annotated. In contrast, we used untrained annotators who were not provided with the document but only with isolated sentences, as in \citet{govindarajan2019decomposing} annotation.

\paragraph{k-Rater Reliability}
We evaluated the reliability of our data as k-rater reliability (kRR), which is a multi-rater generalization of inter-rater reliability (IRR), following the proposal of \citet{wong2022k}. They argue that when aggregate ratings are used as final values, as in our case, k-rater reliability should be used as the correct data reliability. They point out that IRR reports the reliability of the labeling process, while kRR quantifies the reliability of the aggregated data we consume; thus, this seems to be the correct way to account for the reliability of the final data itself.

To compute the kRR we used the Intraclass Correlation Coefficient (ICC), a reliability coefficient for continuous or ordinal rating scales, commonly used in behavioral measurement and psychometrics. ICC gives researchers granular control over assumptions about raters. For example, it is possible to define whether each item is annotated by the same group of raters, or different groups of raters (interchangeability). We used the latter formulation, which allows us to generalize our reliability results to all raters who possess the same characteristics as the raters selected in the reliability study. 

The ICC for k-rater averages is denoted as ICC(k), where \textit{k} stands for the total of the raters (k = 30), following \citet{mcgraw1996forming} notation. For transparency, in Table \ref{tab:tab1} we report both ICC(k) and ICC(1), that is the ICC for the reliability of individual ratings (IRR), emphasising that the first one is the one that account for the reliability of the aggregated data.

\begin{table}[t]
    \centering
    \small
    \scalebox{0.9}{
    \begin{tabular}{ccc}
    \hline
         & ICC(k) & ICC(1) \\
        \hline
    inclusiveness ratings & \textbf{0.97}& 0.52\\
    abstractness ratings & \textbf{0.94} & 0.34\\
    \hline
    \end{tabular}
    }
    \caption{ICC(k) and ICC(1) for inclusiveness and abstractness ratings}
    \label{tab:tab1}
\end{table}

\section{Analysis}
To demonstrate that our continuous scheme subsumes the standard distinction GENERIC vs. NON-GENERIC we compared the aggregated data derived from our annotation with the SitEnt gold annotations. We will refer to the average of inclusiveness ratings as INC and that of abstractness ratings as ABS.

\subsection{Quantitative comparison}

\begin{table}[b]
    \centering
    \small
    \scalebox{0.75}{
    \begin{tabular}{m{2.5cm}cccc}
    \hline
        Predictor(s) & SitEnt label & P & R & F\\
        \hline
        \begin{tabular}[c]{@{}c@{}}\\ INC\\(accuracy = 0.78) \end{tabular}& NON-GENERIC & 0.77 & 0.82 & 0.79\\
        & GENERIC &  0.82 & 0.74 & 0.77 \\
        \hline
        \begin{tabular}[c]{@{}c@{}}\\ ABS\\(accuracy = 0.68) \end{tabular} & NON-GENERIC & 0.65& 0.76& 0.70\\
        & GENERIC &  0.73 & 0.60 & 0.65 \\
        \hline
        \begin{tabular}[c]{@{}c}\\ INC + ABS\\(accuracy = 0.80)\end{tabular} & NON-GENERIC & 0.78&0.83&0.80\\
        & GENERIC &  0.83& 0.76& 0.79 \\
    \hline
    \end{tabular}
    }
    \caption{Prediction of SitEnt labels GENERIC/NON-GENERIC using our continuous annotations in logistic regression models.}
    \label{tab:prediction_model}
\end{table}

 \begin{figure*}[h]
    \centering
    \small
    \includegraphics[width=0.75\textwidth]{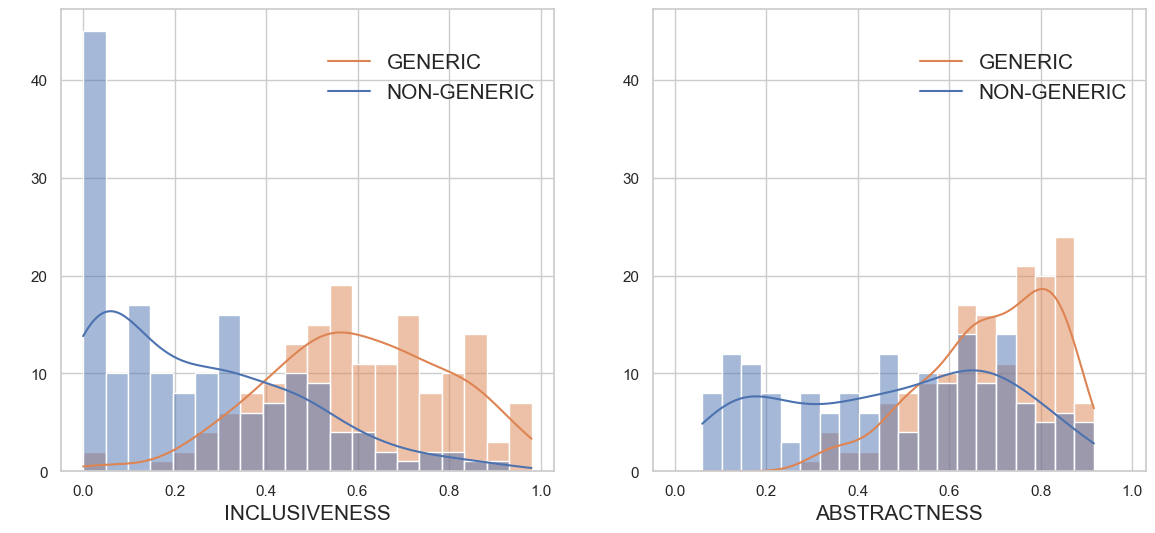}
    \caption{Distribution of INC and ABS by SitEnt labels GENERIC/NON-GENERIC.}
    \label{fig:incl_abstr_dist}
\end{figure*}

\begin{table*}[bp]
    \centering
    \small
    \scalebox{0.9}{
        \begin{tabular}{|m{11cm}|c|c|c|}
        \hline
        \centering{sentence} & INC & ABS & SitEnt label \\
        \hline
        1. \textbf{\underline{Zebras}} evolved among the Old World horses within the last 4 million years. & \multicolumn{1}{|c|}{\cellcolor{orange!97} 0.97} & \cellcolor{orange!88} 0.88 & GENERIC \\
        \hline
        2. He stepped away as the \textbf{\underline{car}} backed up. & \cellcolor{orange!0} 0.00 & \cellcolor{orange!8} 0.08 & NON-GENERIC \\
        \hline
        3. Asian \textbf{\underline{countries}} have tended to give priority to economic, social and cultural rights, but have often failed to provide civil and political rights. & \cellcolor{orange!59} 0.59 & \cellcolor{orange!66} 0.66 & GENERIC \\
        \hline
        4. Many \textbf{\underline{countries}} abolished the monarchy in the 20th century and became republics. & \cellcolor{orange!57} 0.57 & \cellcolor{orange!68} 0.68 & NON-GENERIC \\
        \hline
        5. When the large \textbf{\underline{fish}} of the Colossoma genus entered the aquarium trade in the U.S. and other countries, they were erroneously labeled pacu. & \cellcolor{orange!33} 0.33 & \cellcolor{orange!75} 0.75 & GENERIC \\
        \hline
        6. When a \textbf{\underline{cat}} is dropped, it always lands on its feet. & \cellcolor{orange!92} 0.92 & \cellcolor{orange!53} 0.53 & GENERIC \\
        \hline
        \end{tabular}
        }
        \caption{Some sentences from our dataset; target nouns are underlined.}
    \label{tab:example_sentences}
\end{table*}

Preliminarily, we performed a Wilcoxon rank-sum test to assess whether the difference in INC and ABS between the GENERIC and NON-GENERIC group was statistically significant. The result is highly significant for both INC (\textit{p} = 1.81e-31) and ABS (\textit{p} = 1.14e-15).

Then, we fit three logistic regression models to predict the SitEnt label of each noun on the basis of our annotations: one using only INC as predictor, one using only ABS and one using both.
To select the hyperparameters for these classifiers we used grid search over different solvers (\textit{solver} in \{'newton-cg', 'lbfgs', 'liblinear', 'sag', 'saga'\}) and the regularization parameter C (\textit{C} in \{0.001, 0.01, 0.1, 1, 10\}) in a 5-fold stratified crossvalidation (CV) nested within a 10-fold CV\footnote{To fit the regression models we relied on the scikit-learn Library: \url{https://scikit-learn.org/stable/index}.}. The standard deviation for each metric across the 10 folds is always < 0.15. This means that the performance metrics are stable across folds and that the models are not overly sensitive to the specific subsets of data used for training.
The metrics for each classifier (accuracy; precision, recall and F1-score for each class), computed from this 10-fold CV, are reported in Table \ref{tab:prediction_model}.

The results show good performances of the models, which means that our annotations provided by crowd workers are good predictors for binary classification between the SitEnt labels GENERIC and NON-GENERIC, provided by experts.

Interestingly, all three models perform well, and the best-performing one is that using both INC and ABS as predictors (although the difference is only a few points compared to the model using only INC). This suggests that both semantic dimensions are good predictors for the aspect of genericity, and that the information conveyed by each one adds predictive power to the other.
These results also confirm that the use of untrained annotators, who rely primarily on their intuition, and isolated sentences rather than whole documents, still allows us to capture information about the phenomenon of genericity.

The models perform well enough to confirm that our continuous annotation tracks the binary one. However, the performances are still not excellent, rising the possibility that a binary classification does not reflect the totality of information captured by our continuous annotation.
In fact, the cases of mis-classification GENERIC/NON-GENERIC are probably due to the difference between type of annotation (binary vs. continuous): binary labels force evaluators to one or the other label, and in some cases this choice is likely to be arbitrary (as we will show in the next section and is reported in Table \ref{tab:example_sentences}), while continuous evaluations allow for nuances.

\subsection{Qualitative comparison}
The distribution of INC and ABS in comparison with that of GENERIC/NON-GENERIC labels is shown by the histograms in Figure \ref{fig:incl_abstr_dist}.

The distribution of INC (left) reflects that of the SitEnt labels: nouns labeled as NON-GENERIC cluster on low values of inclusiveness (with a peak on the minimum value, which on our scale corresponds to 'a particular X'), nouns labeled as GENERIC on high values of inclusiveness. This is in agreement with our expectations: in fact, nouns annotated as GENERIC (referring to a kind or an arbitrary member of a kind, according to the SitEnt guidelines) will tend to be perceived as more inclusive, i.e., as referring to more elements of the category denoted by the word, than those annotated as NON-GENERIC (referring to a particular individual, according to the guidelines).
However, the central part of the plot shows a large overlap area, where intermediate values of inclusiveness match both nouns labeled as GENERIC and as NON-GENERIC.

As for ABS (Figure \ref{fig:incl_abstr_dist}, right), nouns labeled as GENERIC tend to cluster on high ratings, while those labeled as NON-GENERIC are fairly evenly distributed along the entire scale. Thus, in this case, the distribution of ratings mirrors that of the binary labels with regard to the GENERIC label, as we expected: nouns referring to a kind tend to be perceived as more abstract. The mixed situation exhibited by nouns labeled as NON-GENERIC, on the other hand, will be at least partly accounted for by the inherent semantics of the noun in question: that is, inherently very abstract nouns, such as \textit{idea}, will tend to always receive high abstractness ratings. We will return to this aspect in §5.3.

The existence of broad areas of overlap between GENERIC and NON-GENERIC nouns in the distribution of INC and ABS confirms that a discrete classification is unable to capture the totality of information conveyed by our continuous annotation. The binary distinction is in full agreement with the continuous annotations as far as they concern the cases of \textit{prototypical} generic and non-generic nouns, which will be indeed characterized by very high values for both INC and ABS (see sentence (1) in Table \ref{tab:example_sentences}) and very low values for both INC and ABS (sentence (2) in Table \ref{tab:example_sentences}), respectively. 

 \begin{figure*}[t]
    \centering
    \includegraphics[width=0.6\textwidth]{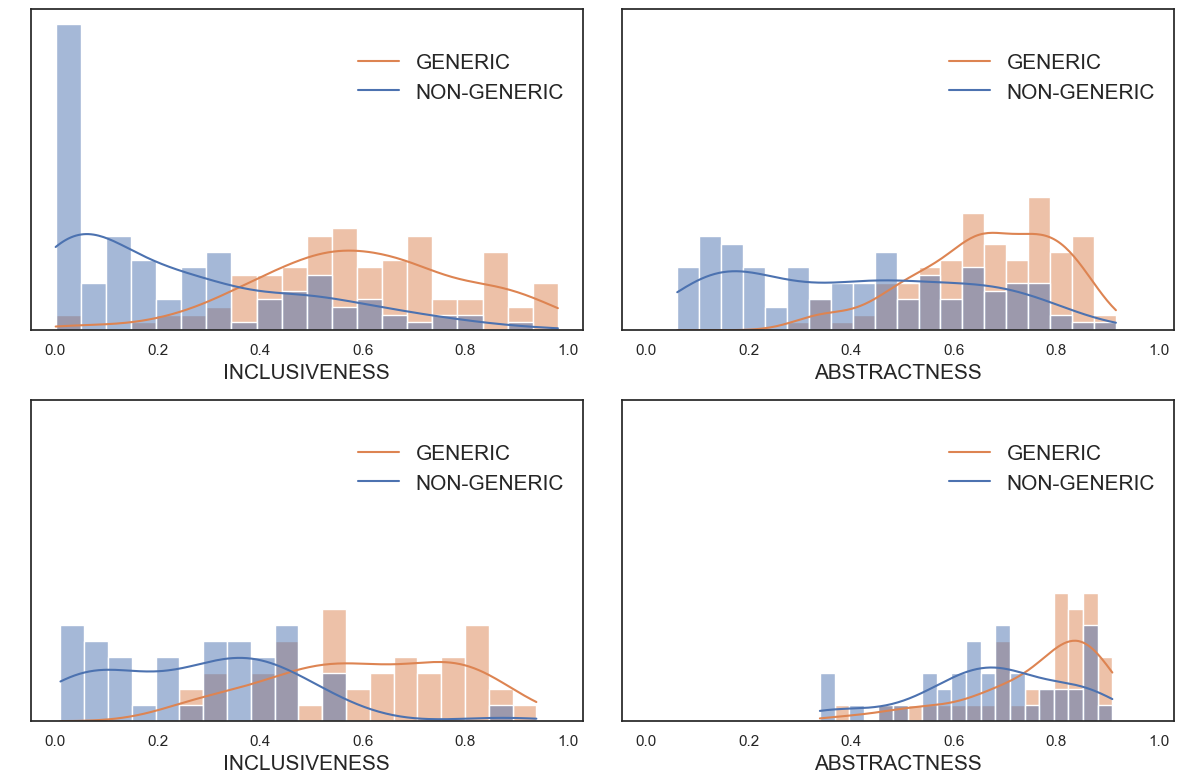}
    \caption{Distribution of INC (left) and ABS (right) for concrete (top) and abstract (bottom) nouns.}
    \label{fig:conr_abst_dist}
\end{figure*}

However, natural language is rarely so clear-cut. Our analysis shows that there are also \textit{non-prototypical} cases that are difficult to classify by binary annotation and that could benefit from a representation through our semantic dimensions. Consider cases (3) and (4) in Table \ref{tab:example_sentences}: in both sentences, the word \textit{countries} does not blatantly refer either to a kind or to a specific individual. This is reflected by intermediate values of INC and ABS, very close for the two cases. However, in SitEnt the first one is classified as GENERIC and the other as NON-GENERIC, which confirms the difficulty of assigning some cases to closed classes.
The last two examples in Table \ref{tab:example_sentences} show that not all generics are the same, and that our annotation can usefully capture fine-grained differences between them. Sentence (5) shows a case of taxonomic reference, where the word \textit{fish} does not refers to the category of all fish, but to a subkind; it is also a case of direct kind predication. These features are well represented by the ratings: the word is low in inclusiveness because refers to a relatively small number of fish, but high in abstractness because it still refers to a kind. In sentence (6), on the contrary, the word \textit{cat} is high in inclusiveness and low in abstractness. This sentence is a characterizing generic, with the subject in its indefinite form: the predication is applicable to a vast majority of individuals of the kind, but the noun, rather than refer directly to the kind, refers to an arbitrary member of the kind, which is perceived as less abstract.

\subsection{Inherent semantics of words}

The last point we consider in our analysis is that of the \textit{inherent} semantics of words evaluated for their in-context genericity, particularly with respect to their concreteness/abstractness. This aspect is rarely addressed explicitly, either by the theoretical literature on genericity or by works on its annotation, probably because the distinction between generic and non-generic meaning is more difficult to draw for nouns referring to abstract entities.

\citet{govindarajan2019decomposing} attempt to address this problem proposing the disentanglement of the following dimensions: KIND, PARTICULAR and ABSTRACT (non-mutually exclusive labels). However, their analysis shows that there is a tendency for abstract-referring nouns to be neither kind-referring nor particular-referring. 
Our claim is that the kind/particular (or generic/non-generic) meaning is well represented by the dimension of inclusiveness for both concrete nouns, as we saw in the previous paragraph, \textit{and} abstract nouns, as can be seen from the difference in INC in a contrast such as:

\begin{enumerate}[resume]
\item
    \begin{enumerate} [noitemsep]
        \item My \textbf{mind} took this as a challenge, something I had to prove wrong. [INC: 0.01; ABS: 0.64]
        \item Substance dualists argue that the \textbf{mind} is an independently existing substance. [INC: 0.87; ABS: 0.90]
    \end{enumerate}
\end{enumerate}

We can rely on the dimension of inclusiveness for these cases as well, without having to take the kind/particular meaning out of the equation for abstract entities.
This can also be inferred from the plots shown in Figure \ref{fig:conr_abst_dist}, analogous to those in Figure \ref{fig:incl_abstr_dist}, but in which the target nouns have been split into concrete (top) and abstract (bottom)\footnote{We recall that we considered as concrete nouns associated with scores $>$ 3 and as abstract nouns associated with scores $\leq$ 3 in \citet{brysbaert2014concreteness}}. 
The distribution of INC (plots on the left) is similar between concrete and abstract nouns: in both cases there is a bimodal trend with respect to binary labels, with the difference that abstracts do not show the same peak as concretes on the minimum value.
\\ \newline
As for ABS, we claim that it is determined both by context \textit{and} by the inherent semantics of the word. The influence of context is particularly evident in the case of concrete entities, as can be seen from the difference in ABS in a contrast such as:

\begin{enumerate}[resume]
\item
    \begin{enumerate} [noitemsep]
        \item The \textbf{cat} sadly shook its head and meowed. [INC: 0.01; ABS: 0.06]
        \item The domestic \textbf{cat} was first classified as Felis catus by Carolus Linnaeus. [INC: 0.85; ABS: 0.89]
    \end{enumerate}
\end{enumerate}

ABS values are also influenced by the inherent semantics of the word, which is particularly evident for nouns referring to abstract entities. In both (4a) and (4b), for example, ABS values are above the midpoint of the scale: this is because the entity in question is inherently abstract (\citealp{brysbaert2014concreteness} score: 2.5). However, interestingly, there is still a difference in ABS between the two cases, which confirms that there is an influence of context in these cases as well.

The influence of the word's inherent semantics thus helps to explain the ABS distribution of NON-GENERIC nouns in Figure \ref{fig:incl_abstr_dist}. This can be better visualized in the plots on the right in Figure \ref{fig:conr_abst_dist}. For concrete nouns (top), the ABS distribution is bimodal between GENERIC and NON-GENERIC, similarly to the INC one. For abstract nouns (bottom), the distribution is skewed toward high values on the scale: in-context abstractness of inherently abstract nouns does not fall below a certain threshold.

\section{Discussion and Conclusions}

In this paper, we introduced a novel annotation framework aimed at capturing the nuances of noun phrases' (NPs) genericity in a fine-grained manner. We discussed our dual objective in proposing this framework: firstly, to examine if naive language users' annotations can reveal differences in genericity that align with phenomena observed in theoretical literature by experts; secondly, we argue that the annotations generated through our framework can serve as valuable training data for systems to automatically identify different levels of genericity, which, in turn, can be employed to construct repositories of commonsense knowledge. 
To validate our annotation scheme, we compared continuous annotations collected in crowdsourcing tasks with existing binary annotations on the same dataset, showing that our continuous annotations reliably capture fine-grained nuances of genericity. 

In conclusion, our work holds potential for linguistics, in that it provides a first dataset of natural occurring sentences annotated according fine-grained continuous values modeling NPs' genericity and an annotation scheme designed to build such annotated datasets, that can be used in semantic studies on genericity. Furthermore, it is also exploitable for the creation of commonsense knowledge repositories, useful for the enhancement of various NLP applications.

\section*{Acknowledgements} 
Funded by the European Research Council, Abstraction project (attributed to Marianna Bolognesi. Grant agreement: ERC-2021-STG-101039777). Views and opinions expressed are however those of the authors only and do not necessarily reflect those of the European Union or the European Research Council Executive Agency. Neither the European Union nor the granting authority can be held responsible for them.

Authors Contribution: CC (conception, experimental design, data collection, data analyses, writing); AAM (conception, revision); MMB (conception, revision, supervision).

We thank the anonymous reviewers for their useful suggestions, and all the members of the research group Abstraction (https://www.abstractionproject.eu/).

\section*{Bibliographical References}\label{sec:reference}
\bibliographystyle{lrec-coling2024-natbib}
\bibliography{lrec-coling2024}

\section{Language Resource References}
\label{lr:ref}
\bibliographystylelanguageresource{lrec-coling2024-natbib}
\bibliographylanguageresource{languageresource}

\end{document}